\pdfoutput=1

\documentclass[11pt]{article}

\usepackage[preprint]{acl}

\usepackage{times}
\usepackage{latexsym}

\usepackage[T1]{fontenc}

\usepackage[utf8]{inputenc}

\usepackage{microtype}

\usepackage{inconsolata}

\usepackage{graphicx}
\graphicspath{ {figures/} }

\usepackage{hyperref}

\usepackage{xcolor}

\usepackage{comment}

\newcommand{\mc}[3]{\multicolumn{#1}{#2}{\textbf{#3}}}

\newcommand{\tb}[1]{\textbf{#1}}
\newcommand{\tr}[1]{{\color{red}\textbf{#1}}}
\newcommand{\by}[1]{{\colorbox{yellow}{#1}}}

\makeatletter
\newcommand{\defeq}{\mathrel{\aban@defeq}}
\newcommand{\aban@defeq}{%
  \vbox{\offinterlineskip\check@mathfonts
    \ialign{\hfil##\hfil\cr
      \fontsize{\ssf@size}{\z@}\normalfont def\cr
      \noalign{\kern1\p@}
      $\m@th=$\cr
      \noalign{\kern-.5\fontdimen22\textfont2}
    }%
  }%
}
\makeatother

\title{A Multiple-Fill-in-the-Blank Exam Approach for Enhancing Zero-Resource Hallucination Detection in Large Language Models}

\author{
  Satoshi Munakata \and Taku Fukui \and Takao Mohri \\
  Fujitsu LTD., Kanagawa, Japan \\
  \texttt{\{munakata.satosi,fukui.taku\}@fujitsu.com}
}

\begin{document}
\maketitle

\begin{abstract}
Large language models (LLMs) often fabricate a hallucinatory text.
Several methods have been developed to detect such text by semantically comparing it with the multiple versions probabilistically regenerated.
However, a significant issue is that if the storyline of each regenerated text changes, the generated texts become incomparable, which worsen detection accuracy.
In this paper, we propose a hallucination detection method that incorporates a multiple-fill-in-the-blank exam approach to address this storyline-changing issue.
First, our method creates a multiple-fill-in-the-blank exam by masking multiple objects from the original text.
Second, prompts an LLM to repeatedly answer this exam.
This approach ensures that the storylines of the exam answers align with the original ones.
Finally, quantifies the degree of hallucination for each original sentence by scoring the exam answers, considering the potential for \emph{hallucination snowballing} within the original text itself.
Experimental results show that our method alone not only outperforms existing methods, but also achieves clearer state-of-the-art performance in the ensembles with existing methods.
\end{abstract}


\section{Introduction}

Generative large language models (LLMs) often fabricate text that contradicts or is not grounded against real-world information.
This harmful phenomenon is known as \emph{Factuality hallucination} (hereinafter simply ``hallucination'') \citep{huang2023survey}.
As LLMs are increasingly adopted for a variety of language-related tasks in daily life and industry, hallucination detection in LLMs is essential to ensure trustworthiness \citep{sun2024trustllm}.

Existing detection methods can be categorized into those that (a) retrieve external facts, (b) analyze LLM's internal state, and (c) use only LLM's input/output (i.e., \emph{zero-resource black-box detection}) \citep{huang2023survey}.
Although each has different pros and cons, this work focuses on type (c), which does not require an external knowledge base and can also apply to LLMs used via only WebAPIs and to domain-specific fine-tuned LLMs.
Among several existing type (c) methods (\citealp{agrawal2023indirect, anonymous2024selfcontradictory, anonymous2024cove, cohen2023lmvslm} as listed in \ref{sec:related_work_methods}), \emph{SelfCheckGPT-Prompt} (hereinafter ``SCGP'') is a reproducible and peer-reviewed state-of-the-art (SOTA) method \citep{manakul-etal-2023-selfcheckgpt}.
SCGP utilizes the nature that hallucinatory text typically exhibits low robustness; i.e., regenerating the consistent text is probabilistically challenging.
Consequently, SCGP uses LLMs to determine whether the original text is semantically supported by each of the probabilistically regenerated texts from the same prompt.
Sentences that lack support are more likely to be considered as hallucinations.

\begin{figure}
  \centering
  \includegraphics[width=7cm]{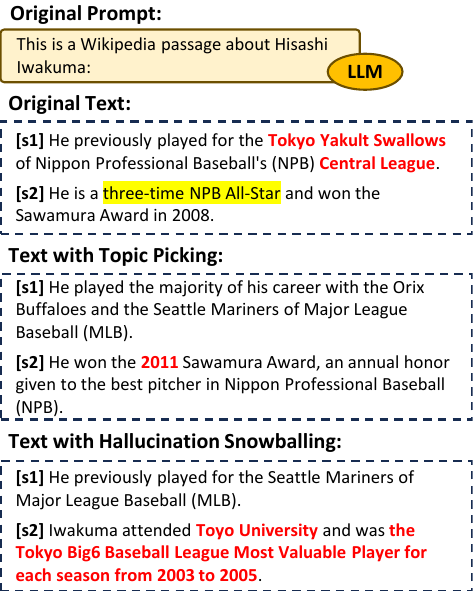}
  \caption{Examples of the storyline-changing issue.
    Each text is generated with the original prompt.
    Each sentence is assigned a serial number, such as \tb{[s1]}.
    \tr{Red bold} indicates hallucinatory phrases.
    \by{Yellow background} indicates non-hallucinatory but incomparable phrases due to the regenerated texts with topic picking and snowballing.
  }
  \label{fig:storyline-changing}
\end{figure}

A significant issue for SCGP is that the storyline of each regenerated text changes, which leads to incomparable sentences in the original text, particularly in the latter part, as exemplified in Figure~\ref{fig:storyline-changing}.
These incomparable sentences worsen detection accuracy because they are determined as hallucinations even when they are not.
The changes in the storyline are not easy to deal with, as they are a mixture of those caused by \emph{topic picking} and \emph{hallucination snowballing} (hereinafter simply ``snowballing'') \citep{zhang2023snowball}.
Snowballing is the phenomenon that LLMs over-commit to early mistakes, which leads to more mistakes that they otherwise would not make.
Contrary to mere topic picking, subsequent sentences in snowballing are highly likely to be hallucinations (cf. \ref{sec:example_snowballing}).
\begin{figure*}
  \centering
  \includegraphics[width=\textwidth]{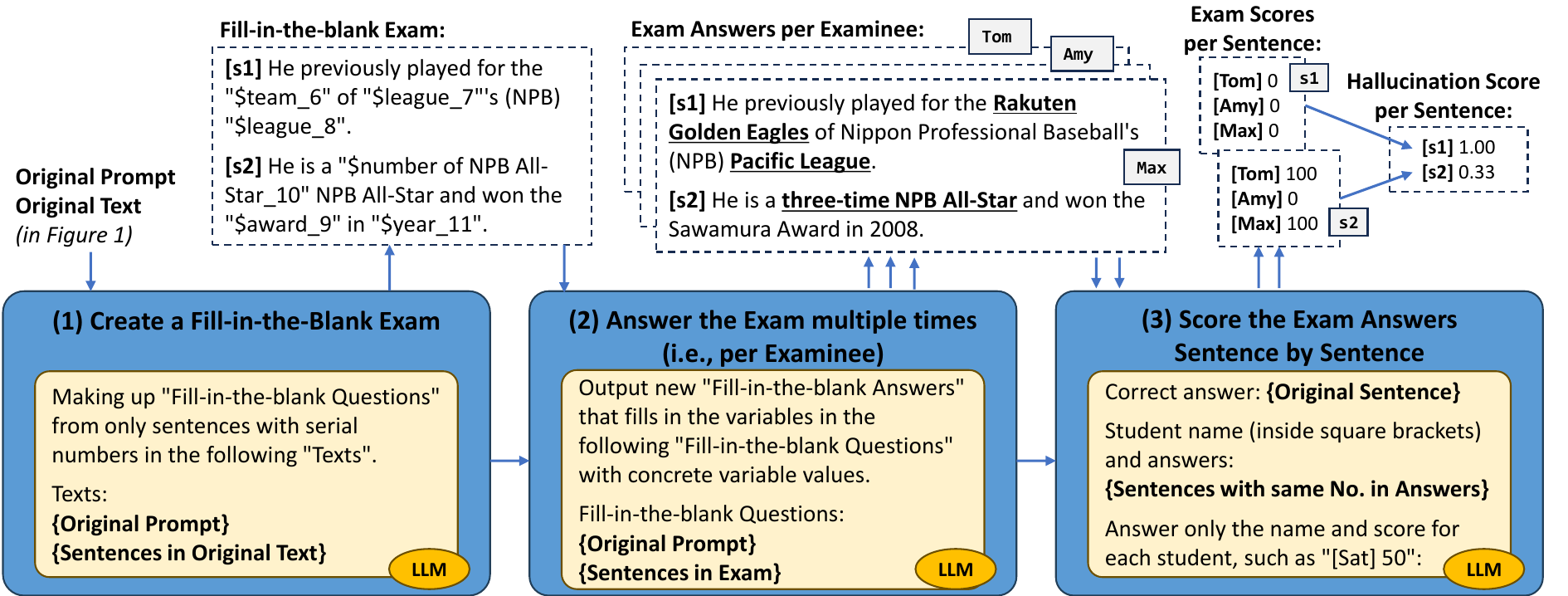}
  \caption{An example of our FIBE approach with the original text in Figure~\ref{fig:storyline-changing}.
    This exemplifies the steps to predict the hallucination score for each sentence in the original text.
    \underline{\tb{Bold underline}} in the exam answers indicates comparable phrases that were regenerated according to our expectations and that correspond to the \tr{hallucinatory} or \by{incomparable} phrases in Figure~\ref{fig:storyline-changing}.
  }
  \label{fig:example_FIBE}
\end{figure*}

In this paper, we propose a novel zero-resource hallucination detection method that incorporates a \emph{multiple-fill-in-the-blank exam (FIBE)} approach for the above storyline-changing issue.
Figure~\ref{fig:example_FIBE} shows an example of our FIBE approach.
First, instead of merely regenerating, (1) creates a multiple-fill-in-the-blank exam by masking multiple objects from the original text.
Second, (2) prompts an LLM to repeatedly answer this exam with some additional hints.
This approach ensures that the storylines of the exam answers align with the original one, thereby preventing the emergence of incomparable sentences.
Finally, (3) quantifies the degree of hallucination for each original sentence by scoring the exam answers.
In this scoring, considering the potential for snowballing within the original text itself, we further propose to use 2 approaches; \emph{Direct Question (DQ)} and \emph{Snowballing Correction (SBC)}.
%
We compare the performance of our method with the existing method SCGP using \emph{the WikiBio GPT-3 Hallucination Dataset v3} \citep{wikibiogpt3hallucination}.

\tb{Main Contributions:}
\tb{(i)} We proposed a novel hallucination detection method incorporating our \emph{FIBE}, \emph{DQ} and \emph{SBC} approaches that enable more precise comparative analysis against the storyline-changing issue involving topic picking and snowballing. This method achieved SOTA detection accuracy.
\tb{(ii)} We discovered a decline in detection accuracy in multi-line LLM-generated text, particularly noticeable from the second line onward, for the first time. By addressing this issue, both our method alone and in an ensemble with SCGP show a clear improvement in accuracy over SCGP alone.

\section{Notation}

$r_i$ is the $i$-th sentence in original LLM response text $R$ generated from prompt $P$.
A hallucination detection method $H$ predicts hallucination score $H(i) \in [0,1]$ of $r_i$.
Ideally, the more hallucinatory $r_i$ is, the higher $H(i)$ should be.
Variants are distinguished by the subscript of $H$.
%
For example, existing method SCGP is denoted as $H_{P}(i) \defeq N^{-1} \Sigma^N_j (1 - supported(r_i, sample^j(P)))$; where $sample^j(P) = S^j$ is the $j$-th probabilistically regenerated text from prompt $P$, $N$ is the maximum sample count, and $supported(r_i, S^j) \in [0,1]$ is the value so high that $r_i$ is supported by text $S^j$.
Function $supported$ is realized with the LLM prompt in \ref{sec:prompt_SCGP}.

\section{Methodology}

In the SCGP, if the storyline-changing occurs in regenerated text $S^j$ and sentence $r_i$ is no longer comparable to $S^j$ as in Figure~\ref{fig:storyline-changing}, $H_{P}(i)$ predicts that $r_i$ is hallucinatory even if it is not because $supported(r_i, S^j)$ can only take a low value.
Accordingly, we propose \emph{FIBE} approach to forcefully regenerate comparable sentences with each $r_i$.
Furthermore, we propose \emph{DQ} and \emph{SBC} approaches to consider \emph{snowballing} that occurred within original text $R$ itself.

\subsection{Fill-in-the-Blank Exam (FIBE)}

As shown in Figure~\ref{fig:example_FIBE}, FIBE regenerates sentences that match other constructions, such as subjects and verbs, by creating a multiple-fill-in-the-blank exam with multiple objects masked in original text $R$ and prompting an LLM to repeatedly answer it.
Here, the objects appearing before the subject in each sentence are not masked to prevent topic picking.
%
FIBE is denoted as $H_{F} \defeq 1 - (100N)^{-1} \Sigma^N_j\\score(answer^j_i(create(R, P), P), r_i)$; where $create(R, P) = E$ is the exam based on text $R$ under the context of $P$, $answer^j_i(E, P) = a^j_i$ is the $i$-th sentence of the $j$-th answer for exam $E$ under the context of $P$, and $score(a^j_i, r_i) \in [0, 100]$ is the value so high that answer $a^j_i$ is consistent with $r_i$.
Functions $create$, $answer$, and $score$ are realized with the LLM prompts in \ref{sec:prompt_FIBE_create}, \ref{sec:prompt_FIBE_answer}, and \ref{sec:prompt_FIBE_score}, respectively.
Here, SCGP's $supported(r_i, S^j)$ compares a sentence with a text, whereas FIBE forcefully obtains comparable sentence $a^j_i$, so that $score(a^j_i, r_i)$ can compare a sentence with a sentence. This considerably reduces the size of prompt tokens.

\subsection{Direct Question (DQ)}

If snowballing occurs in original sentence $r_i$, and if it occurs in the exam answer $a^j_i$ as well, $score(a^j_i, r_i)$ predicts that $r_i$ is fact.
Therefore, DQ prompts the LLM to answer directly whether original sentence $r_i$ is hallucinatory or not, excluding the influence of the preceding sentences $r_{<i}$.
DQ is denoted as $H_{D}(i) \defeq 1 - known(r_i, P)$; where $known(r_i, P)$ is the value so high that the LLM is convinced that $r_i$ is fact based on its prior knowledge under the context of $P$.
Function $known$ is realized with the LLM prompt in \ref{sec:prompt_DQ}.

\subsection{Snowballing Correction (SBC)}

If snowballing occurs in original text $R$, the more its former sentences are hallucinatory, the more likely the latter sentences are also hallucinatory.
Therefore, in SBC, the hallucination scores in the former part add up to the latter part.
SBC is denoted as $H_{S}(i; H, \theta) \defeq clip(H(i) + |R|^{-1} max(0, \Sigma^{i-1}_{k=0} H(k) - \theta))$; where $H$ is arbitrary detection method, $|R|$ is the number of sentences in $R$, $\theta$ is a constant hyperparameter for adjusting the effectiveness of this correction, and $clip(n)$ is the function to round $n$ in $[0,1]$.

\subsection{Ensembles}

We define the ensemble of multiple detection methods other than SBC as a clipped weighted sum; i.e., $H_{+}(i) \defeq clip(C_{F} H_{F}(i) + C_{D} H_{D}(i) + C_{P} H_{P}(i))$; where $C_F$, $C_D$, and $C_P$ are the constant weights that are hyperparameters.
We also define the ensemble with SBC as a function composite; i.e., $H_{\circ}(i; \theta) \defeq H_{S}(i; H_{+}, \theta)$.

\section{Experimental Evaluation}

\begin{table*}
  \centering
  \caption{Benchmark result. 
    Numbers in \tb{bold} indicate superiority over both \emph{SCGP+ (original)} and \emph{SCGP* (resampled)}.
    Numbers in \tr{red bold} indicate the best value in the same indicator (column).
  }
  \small
  \begin{tabular}{|ll|ccc|cc|}
    \hline
    \mc{2}{|c|}{}                      & \mc{3}{c|}{AUC-PR [\%]}                                        & \mc{2}{c|}{AUC-ROC [\%]}           \\
    \mc{2}{|c|}{Method}                & \mc{1}{c}{NonFact} & \mc{1}{c}{NonFact*} & \mc{1}{c|}{Factual} & \begin{tabular}{c}\tb{NonFact}\\\tb{(Factual)}\end{tabular}
                                                                                                                      & \mc{1}{c|}{NonFact*} \\
    \hline\hline                                                                                                                                    
    Baseline    & SCGP+ (original)     &     91.47          &     61.92           &     64.51           &     78.91   &     68.25            \\
                & SCGP* (resampled)    &     91.55          &     67.53           &     67.26           &     77.88   &     70.72            \\
    \hline                                                                                                                                         
    Ours        & FIBE                 & \tb{91.72}         & \tb{67.54}          &     66.40           & \tb{81.06}  & \tb{71.09}           \\
                & FIBE, DQ             & \tb{92.04}         & \tb{68.40}          & \tb{68.70}          & \tb{81.99}  & \tb{72.04}           \\
                & FIBE, SBC            & \tb{92.77}         & \tb{71.86}          & \tb{70.02}          & \tb{82.89}  & \tb{73.20}           \\
                & FIBE, SBC, DQ        & \tb{92.82}         & \tb{72.66}          & \tb{71.25}          & \tb{82.90}  & \tb{73.55}           \\
    \hline                                                                                                                             
    Ensemble    & FIBE, SCGP*, SBC     & \tr{94.41}         & \tb{73.31}          & \tb{75.45}          & \tr{87.15}  & \tb{77.99}           \\
    with FIBE   & FIBE, SCGP*, SBC, DQ & \tb{94.34}         & \tr{74.25}          & \tr{75.81}          & \tb{86.93}  & \tr{78.04}           \\
    \hline                                                                                                                                       
    Ensemble    & SCGP*, DQ            & \tb{92.00}         & \tb{68.28}          &     60.77           & \tb{81.03}  & \tb{72.76}           \\
    w/o FIBE    & SCGP*, SBC           & \tb{92.78}         & \tb{70.42}          &     66.97           & \tb{82.50}  & \tb{73.94}           \\
    (for refs.) & SCGP*, SBC, DQ       & \tb{92.96}         & \tb{70.89}          &     65.75           & \tb{83.11}  & \tb{74.17}           \\
    \hline
  \end{tabular}
  \label{tab:benchmark_result}
\end{table*}


\subsection{Experimental Details} \label{sec:experimental_details}

\tb{Dataset:}
We used \emph{the WikiBio GPT-3 Hallucination Dataset v3} \citep{wikibiogpt3hallucination} for evaluating zero-resource black-box detection methods.
This dataset originally provides a total of 1,908 sentences in 238 original texts generated by \emph{GPT-3 (text-davinci-003)} using the prompt template \textit{``This is a Wikipedia passage about \{concept\}:''}; where the placeholder \textit{concept} is replaced by one out of 238 person names.
However, we excluded 2 texts because their sentences were originally misdivided in the middle of proper nouns (cf. \ref{sec:incorrect_separated_sentences}).
Thus, 1,893 sentences of 236 texts were evaluated in this experiment.
Each sentence is manually annotated with 3 levels of hallucination intensity; \emph{Major Inaccurate}, \emph{Minor Inaccurate}, and \emph{Accurate}.
This dataset also provides probabilistically regenerated texts using the same GPT-3.

\tb{Tasks and Indicators:}
We evaluated each method on 3 tasks, \emph{NonFact}, \emph{NonFact*}, and \emph{Factual}, which involved binary classification of each sentence in the original texts.
NonFact is the task to classify Major/Minor Inaccurate and others, NonFact* is for Major Inaccurate and others, and Factual is for Accurate and others.
Then, we quantified the single run accuracy of each task using \emph{AUC-PR} and \emph{AUC-ROC} (cf. \ref{sec:evaluation_indicators}).
Note that the AUC-ROC of the NonFact and Factual are always the same.

\tb{Baselines:}
We employed \emph{gpt-3.5-turbo-16k-0613}, the stable version of \emph{OpenAI GPT-3.5} \citep{openai2022chatgpt} at the time of this experiment, as the LLM used by our method and SCGP.
GPT-3.5 is the model used by SCGP when it achieved the highest accuracy in \citep{manakul-etal-2023-selfcheckgpt}.
We evaluated SCGP with the regenerated 5 texts originally provided by the dataset (named \emph{SCGP+}), and SCGP with the new regenerated 5 texts using GPT-3.5 and the prompt in \ref{sec:prompt_resample} (named \emph{SCGP*}).
This is because our method used the same GPT-3.5 to regenerate 5 texts (i.e., answer an exam 5 times), for a fairer comparison.
We also evaluated our method and several ensembles with fixed hyperparameters; $N = 5$, $\theta = 0.1$, $C_D = 0.2$, and $C_F, C_P = 0.5$ if both FIBE and SCGP* are used, otherwise 1.0 for the one used and 0.0 for the one not used.

\subsection{Experimental Result}

\tb{RQ1: \textit{Does the proposed method outperform the existing method SCGP in detection accuracy?}}
Table~\ref{tab:benchmark_result} shows the all indicators of the evaluated tasks.
FIBE alone is inferior to SCGP* in only Factual AUC-PR, but superior to both SCGP+ and SCGP* in all 5 indicators when combined FIBE with DQ or/and SBC.
In contrast, the Factual AUC-PR of SCGP* is rather degraded when combined with DQ or/and SBC.
Therefore, DQ and SBC are complementary approaches to FIBE.
The ensemble of FIBE and SCGP* is the highest in all 5 indicators, that is they are also complementary.

\tb{RQ2: \textit{What factors make the proposed method and the ensemble outperform the SCGP?}}
\begin{figure}
  \centering
  \includegraphics[width=6.5cm]{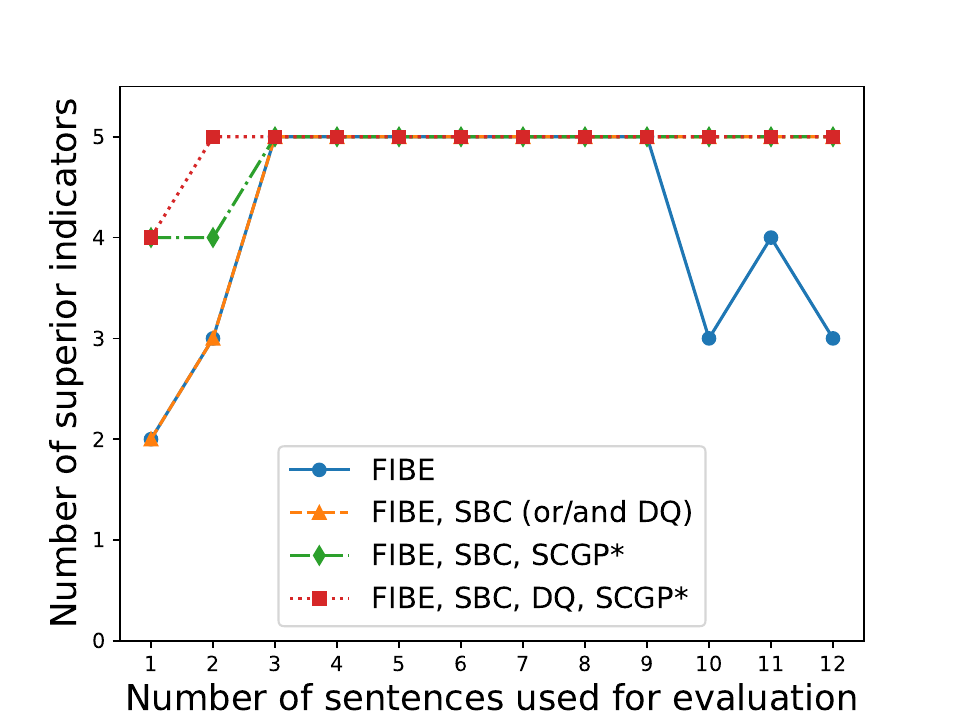}
  \caption{Number of indicators that outperform \emph{SCGP*} (resampled) when the 5 indicators in Table~\ref{tab:benchmark_result} are evaluated using only the first to $x$-th line of each text.}
  \label{fig:wins_SCGP-35}
\end{figure}
Figure~\ref{fig:wins_SCGP-35} shows that each method has different sentence positions in which it excels.
FIBE alone outperforms SCGP* in all 5 indicators when just only classifying from the first to the middle sentences.
This indication supports our hypothesis that the SCGP accuracy tends to be degraded due to the storyline-changing during text regeneration.
The combined use of DQ or/and SBC has the effect of improving accuracy for FIBE when classifying from the first to the last sentences.
%
This indication supports our hypothesis that the FIBE accuracy tends to be degraded due to the snowballing during original text generation; 
i.e., if snowballing produces an irrelevant sentence in the original text's latter half, FIBE's \textit{``forcing comparable sentences''} action is ineffective.
%
Finally, the combined use of SCGP* has improved accuracy in the first sentences. This is also the factor of the outperformance.

\section{Conclusion}

\emph{FIBE}, \emph{DQ} and \emph{SBC} approaches, that we propose in this paper for zero-resource hallucination detection, enable more precise comparative analysis against the storyline-changing issue.
We encourage future work to evaluate them in more diverse LLM use cases; e.g., \emph{RAG} \citep{gao2024rag}.

\newpage

\section{Limitations}

\subsection{Evaluation Indicators} \label{sec:evaluation_indicators}

We omitted the use of any passage-level indicators used by \emph{SelfCheckGPT} \citep{manakul-etal-2023-selfcheckgpt}. 
Because, the order of their values was completely consistent with the order of sentence-level indicator \emph{AUC-PR}.
%
By contrast, we added AUC-ROC into our indicators, which differs from AUC-PR in its curve shape trade-off (cf. \ref{sec:detailed_evaluation_result}).
Because, we found that AUC-PR becomes too high when the number of unique observed values is low like SCGP.

\subsection{Diversity of Experiments}

This work lacks the diversity of benchmark datasets and LLMs.
\emph{The WikiBio GPT-3 Hallucination Dataset v3} \citep{wikibiogpt3hallucination} we used contains only 238 English texts like Wikipedia biographic articles, which are generated from the same prompt template.
Therefore, we should evaluate the external validity of our method using more diverse prompts and topics; e.g., using \emph{the PopQA dataset} \citep{mallen-etal-2023-trust}.
%
Although we used only \emph{GPT-3.5} \citep{openai2022chatgpt} as LLMs in this work, the architecture of our method is not limited to GPT-3.5.
Therefore, we should assess the accuracy when using each of the commonly available LLMs; e.g., \emph{Llama 2} \citep{touvron2023llama2}.

\subsection{Hallucinations in our method itself}

We should investigate the impact of the hallucinations created by our method itself.
In particular, the hallucinatory number generated from quantification prompt $score$ has a direct impact on accuracy.
Increasing the number of resampled texts can be expected to mitigate such impacts.
This work was done with 5 samples and \emph{SelfCheckGPT} done with a maximum of 20 samples.
Furthermore, we should investigate the impact of the stochastic fluctuations of LLM output in our method.
The random seed was fixed to 0 when our method used GPT-3.5 in this work, and the minor version of GPT-3.5 was fixed at \emph{gpt-3.5-turbo-16k-0613} (cf. \ref{sec:complete_prompts}).
However, in order to assess the robustness of differences in random seeds, we should quantify the multiple run accuracy using multiple random seeds.

\subsection{Mathematical Theories}

This work lacks any mathematical theories.
We just use prompt $score$ in \ref{sec:prompt_FIBE_score} to make an LLM compare exam sentences with the original sentence. 
Of course, we also tried many scoring approaches; e.g., to compare named entities using embeddings vector similarity, to compare atomic claims by Chain-of-Thought prompting, etc.
However, this simple $score$ was the most stable and accurate.

\subsection{Hyperparameter/Prompt Tuning}

The fixed common hyperparameters for our experiment listed in \ref{sec:experimental_details} were determined empirically, not optimized for each baseline.
In particular, the fixed weights $C_F$ and $C_P$ when ensembling FIBE and SCGP were set to 0.5, which takes a simple average to eliminate arbitrariness.
However, the possibility of overfitting to a specific baseline/benchmark cannot be ruled out from the experimental result with only one dataset in this paper alone.
Also, we should investigate the impact of the LLM parameters for each prompt.
In this work, we used different \emph{temperature} and \emph{top\_p} parameters of GPT-3.5 for each prompt in order to stabilize the instruction-following results (cf. \ref{sec:complete_prompts}).
%
The $create$ and $answer$ prompts contain one-shot for exemplification of input/output formats (cf. \ref{sec:prompt_FIBE_create} and \ref{sec:prompt_FIBE_answer}).
There is also the possibility that the one-shot is overfitting.

\subsection{Performance Evaluation}

As this paper focuses on accuracy, performance evaluation is lacking.
Nevertheless, this work only used GPT-3.5 via Web API, so few computational resources are required.
FIBE basically has a longer waiting time than SCGP due to the time required to create an exam.
By contrast, FIBE consumes fewer tokens than SCGP because prompt $score(a^j_i, r_i)$ does not require whole regenerated text $S^j$, unlike prompt $supported(r_i, S^j)$.
FIBE requires $1+N+|R|$ times LLM completions per original text, DQ for $|R|$, and SCGP for $N+N|R|$ times; where $N$ is the number of text regenerations and $|R|$ is the number of original sentences.

\section{Ethics Statement}

We acknowledge and ensure that this work is compatible with \emph{the ACL Code of Ethics}.
We note that if hallucinatory sentences are not detected, it could lead to misinformation.
%
\emph{The WikiBio GPT-3 Hallucination Dataset v3} we used is available on \emph{Hugging Face} under the \emph{CC-BY-SA-3.0} license \citep{wikibiogpt3hallucination}.
Our first author manually checked all 238 people who were the topic of each article in this dataset to ensure that they were well-known persons who did not need to be anonymized.

We used AI assistant \emph{GPT-4} \citep{openai2022chatgpt} to check the English grammar of this paper.


\newpage
\bibliography{custom}

\appendix

\section{Detailed Related Work} \label{sec:related_work}

This section describes the relevance of existing studies that have not been discussed so far.

\subsection{LLM Hallucinations} \label{sec:related_work_types}
In addition to Factuality hallucination, which is the main target of this work, there are various other types of hallucinations. 
\emph{Faithfulness hallucination} means that LLM's output is inconsistent with the prompt or intermediate outputs, also known as \emph{Intrinsic hallucination} \citep{huang2023survey}.
\emph{Extrinsic hallucination} means that LLM's output is unverifiable from the prompt \citep{cao-etal-2022-hallucinated}.
These hallucinations can be detected directly by matching prompts and (intermediate) outputs, as in \citep{adlakha2023gpt4eval} and \citep{anonymous2024selfcheck}.
Although our method does not directly support these hallucinations, if they exhibit low robustness like Factuality hallucination, our method can consequently detect them.

\subsection{Zero-Resource Black-box Hallucination Detection} \label{sec:related_work_methods}
Several zero-resource black-box hallucination detection methods have been proposed since SelfCheckGPT was published; however, many of them were under peer review during this work.

SCGP is the last variant added to the \emph{SelfCheckGPT} series.
Because the SCGP performs better than any other variants \citep{manakul-etal-2023-selfcheckgpt}, we did not conduct experiments on the other variants.
\emph{Self-Contradictory} \citep{anonymous2024selfcontradictory} can be regarded as a ``single''-fill-in-the-blank approach and is expected to mitigate the effects of topic picking to some extent; however, there are no approaches against snowballing in the original text.
In comparison with only figures reported in existing papers, the ensemble \emph{``FIBE, SBC, DQ''} outperforms the \emph{Self-Contradictory} in \citep{anonymous2024selfcontradictory}, and even \emph{``FIBE, SCGP*, SBC (, DQ)''} also outperforms the \emph{WikiBio+Prompt} (this is not a zero-resource method because it uses external knowledge) in \citep{manakul-etal-2023-selfcheckgpt}.

Direct Query in \citep{agrawal2023indirect} is similar to our DQ in that it directly asks the LLM for the validity of a single sentence (precisely one bibliography); however differs in that it also refers to the original prompt to spot snowballing.

Coordinating multiple types of LLMs \citep{cohen2023lmvslm} and Chain-of-thought prompt engineering specializing in hallucination detection \citep{anonymous2024cove} are interesting directions and will be future work.

\section{Implementaion Details}
We implemented the proposed method as a Python \footnote{\url{https://www.python.org/}} tool.
The OSS \emph{scikit-learn} \citep{scikit-learn} was used to calculate the AUC values for each of the evaluation indicators.

\section{Detailed Evaluation Result} \label{sec:detailed_evaluation_result}

\begin{figure}[p]
  \centering
  \includegraphics[width=\linewidth]{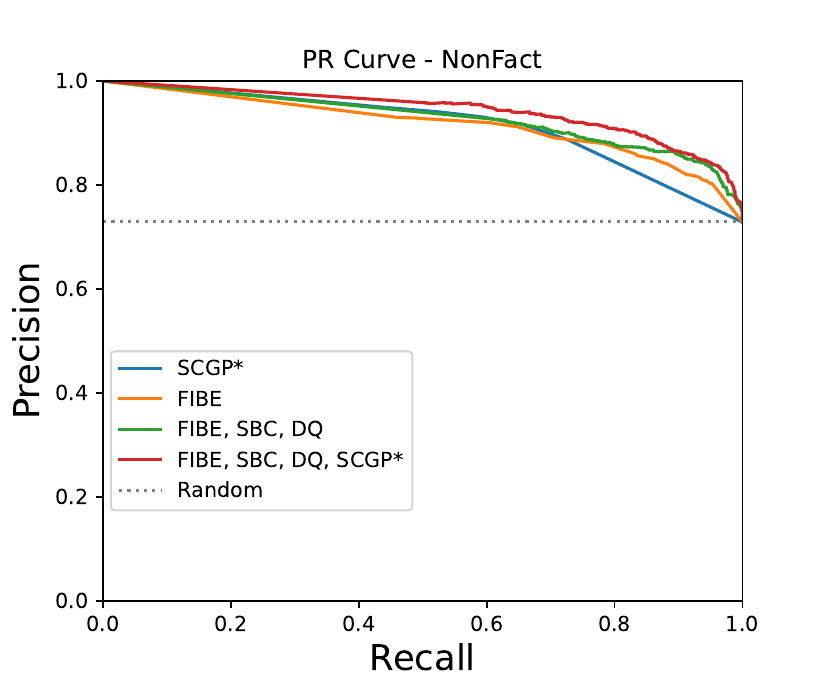}
  \caption{PR Curve - NonFact task}
  \label{fig:PR_NonFact}
\end{figure}

\begin{figure}
  \centering
  \includegraphics[width=\linewidth]{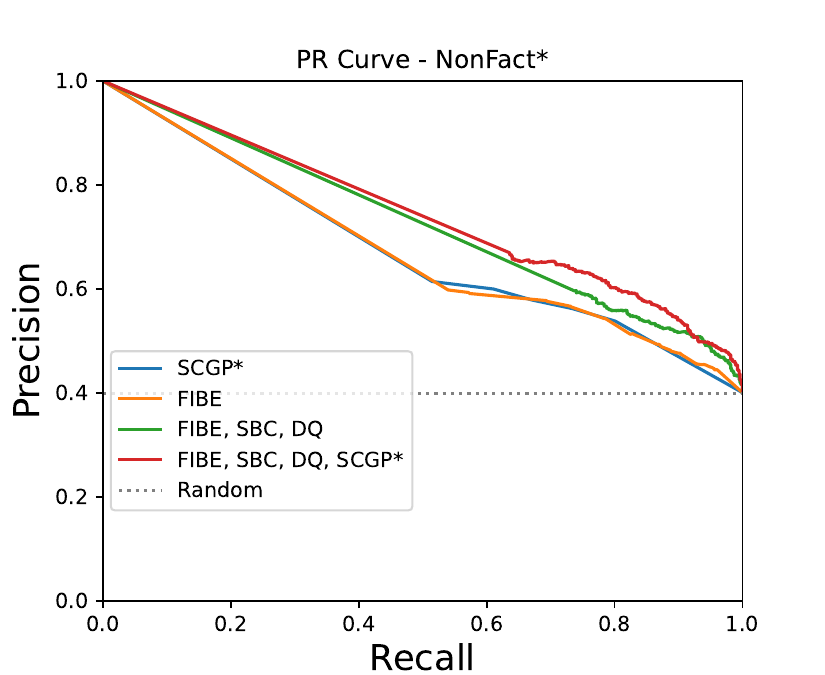}
  \caption{PR Curve - NonFact* task}
  \label{fig:PR_NonFact_}
\end{figure}

\begin{figure}
  \centering
  \includegraphics[width=\linewidth]{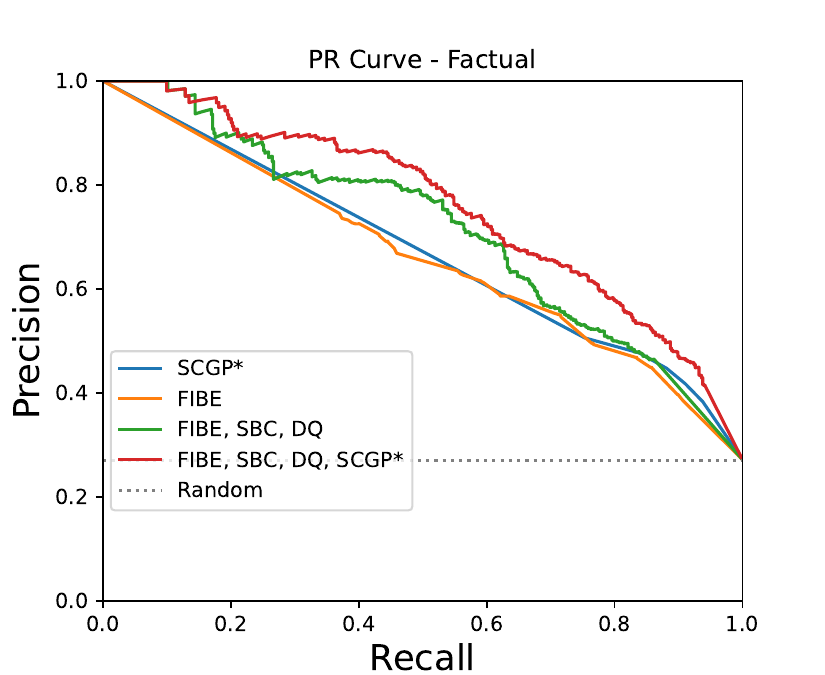}
  \caption{PR Curve - Factual task}
  \label{fig:PR_Factual}
\end{figure}

\begin{figure}
  \centering
  \includegraphics[width=\linewidth]{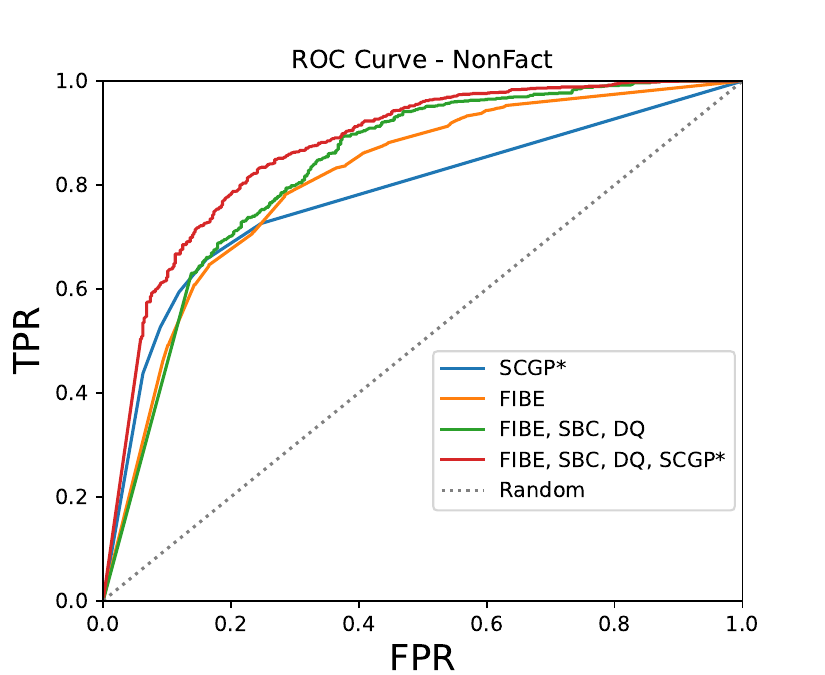}
  \caption{ROC Curve - NonFact task}
  \label{fig:ROC_NonFact}
\end{figure}

\begin{figure}
  \centering
  \includegraphics[width=\linewidth]{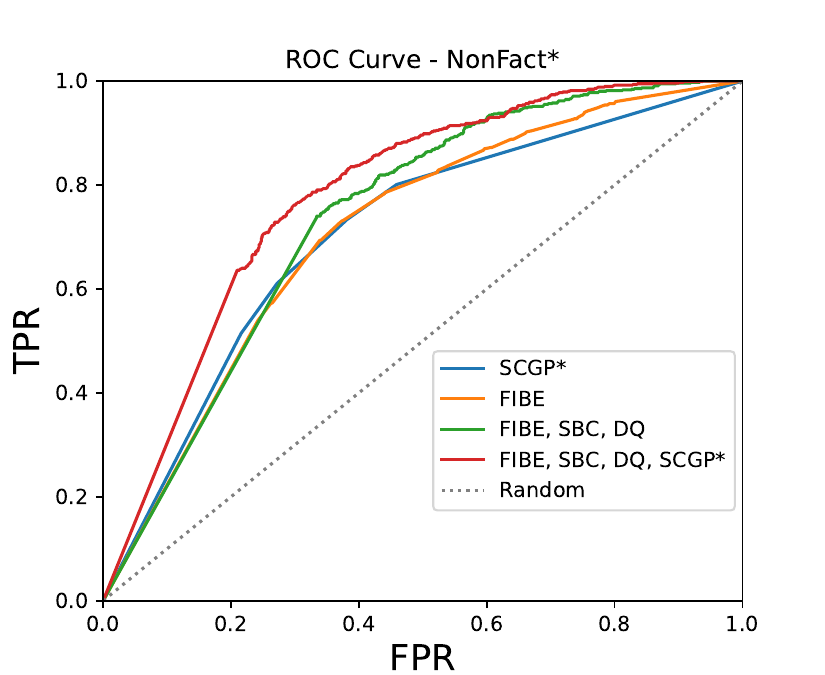}
  \caption{ROC Curve - NonFact* task}
  \label{fig:ROC_NonFact_}
\end{figure}

\begin{figure}
  \centering
  \includegraphics[width=\linewidth]{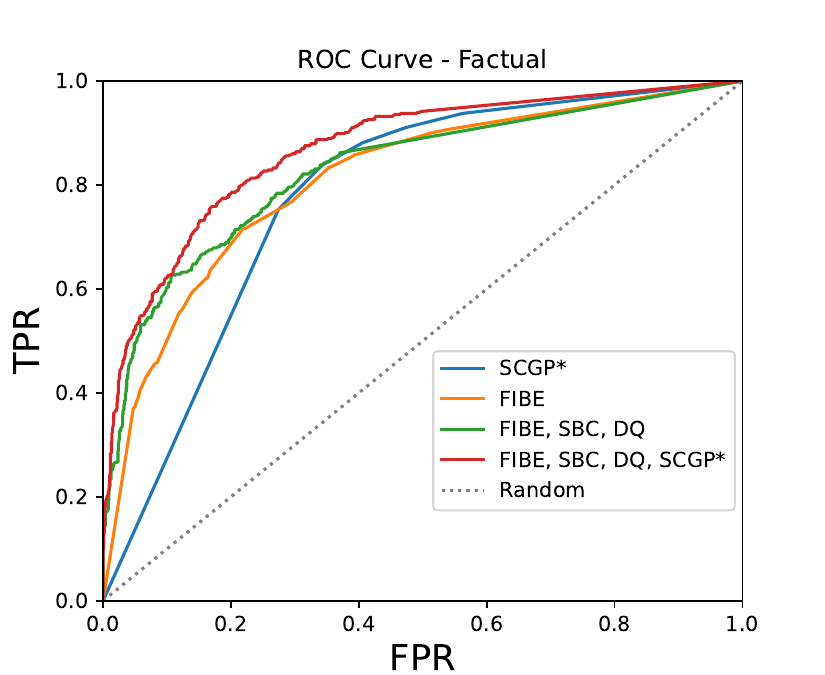}
  \caption{ROC Curve - Factual task}
  \label{fig:ROC_Factual}
\end{figure}

Of our experimental result, the PR and ROC curves for NonFact, NonFact*, and Factual tasks are shown in Figures~\ref{fig:PR_NonFact}, \ref{fig:PR_NonFact_}, \ref{fig:PR_Factual}, \ref{fig:ROC_NonFact}, \ref{fig:ROC_NonFact_}, and \ref{fig:ROC_Factual}, respectively.


\onecolumn
\section{Originally Misdivided Sentences} \label{sec:incorrect_separated_sentences}

\begin{figure}[h]
  \centering
  \includegraphics[width=\linewidth]{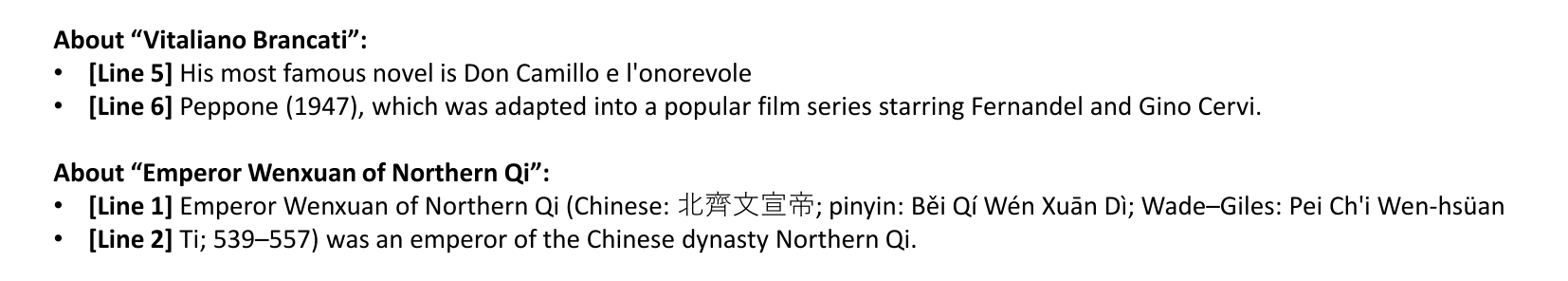}
  \caption{Originally misdivided sentences we excluded in our experiment.
    The text about \textit{``Vitaliano Brancati''} was misdivided in the middle of his novel name.
    The text about \textit{``Emperor Wenxuan of Northern Qi''} was misdivided in the middle of his Wade–Giles style name.
  }
  \label{fig:example_misdivided}
\end{figure}

Figure~\ref{fig:example_misdivided} shows the sentences we excluded in our experiment due to originally misdivided in the middle of proper nouns.
Of course, the same original sentences can be found directly in \emph{the WikiBio GPT-3 Hallucination Dataset v3} \citep{wikibiogpt3hallucination}.
We need to carefully consider how to handle the sentence-level hallucination evaluation of such misdivided sentences.

\section{Complete Prompts} \label{sec:complete_prompts}

This section describes the actual prompt templates used in our experiment and examples of their executions for passage No.28,011 (about \textit{``Bryan McClendon''}).

\tb{Note that:}
\begin{itemize}
  \item \tr{The names of persons and institutions and their relationships exemplified in the following prompts may not be true}
  \item \tr{Expressions that allow one to guess the author (us) are anonymized}
  \item Line breaks are inserted into each example as needed due to space limitations
  \item Some special characters are replaced with one-byte symbols for display/printing purposes
  \item ``\verb|---------------------------------------------------------------|'' in each prompt denotes a role switch \footnote{\url{https://platform.openai.com/docs/api-reference/chat}}
  \item ``\verb|{...}|'' in each prompt denotes a placeholder
\end{itemize}

\subsection{Resampling Dataset} \label{sec:prompt_resample}

The LLM prompt was used for resampling the comparison texts in \emph{the WikiBio GPT-3 Hallucination Dataset v3} \citep{wikibiogpt3hallucination} with \emph{GPT-3.5} \citep{openai2022chatgpt}.

\tb{LLM parameters:}

\begin{itemize}
  \item model version = \emph{gpt-3.5-turbo-16k-0613}
  \item temperature = 1.0 (same value as \citep{wikibiogpt3hallucination})
  \item top\_p = 1.0
  \item random seed = 0
\end{itemize}

\tb{Prompt Template:}

\begin{quote}
\begin{verbatim}
You are the author of an excellent Wikipedia article. 
Write an article of about {n_sentences} lines.
---------------------------------------------------------------
This is a Wikipedia passage about {concept}:
\end{verbatim}
\end{quote}

Where,
\begin{itemize}
  \item \verb|{n_sentences}| is replaced by the number of sentences in the original text
  \item \verb|{concept}| is replaced by one out of 238 person names; such as \textit{``Bryan McClendon''}
\end{itemize}

\tb{Prompt Example:}

\begin{quote}
\begin{verbatim}
You are the author of an excellent Wikipedia article. 
Write an article of about 11 lines.
---------------------------------------------------------------
This is a Wikipedia passage about Bryan McClendon:
\end{verbatim}
\end{quote}

\tb{Completion Example:}

\begin{quote}
\begin{verbatim}
Bryan McClendon is an accomplished American football coach and former player. Born 
on May 27, 1984, in Atlanta, Georgia, McClendon has an impressive career in the 
sport. He played as a wide receiver for the University of Georgia Bulldogs from 
2002 to 2005, earning recognition for his exceptional skills and leadership on the 
field.

Following his playing career, McClendon embarked on a successful coaching journey. 
He held various coaching positions, including serving as the wide receivers coach 
for the University of Georgia from 2009 to 2015. During this time, he played a 
crucial role in developing talented receivers and contributing to the team's 
success.

In 2015, McClendon was named the interim head coach for the Georgia 
Bulldogs, becoming the youngest head coach in the Southeastern Conference at just 
31 years old. Despite only leading the team for one game, he showcased his ability 
to inspire and motivate the players.

McClendon's coaching prowess caught the attention of the National Football League 
(NFL) teams, and in 2016, he joined the staff of the Carolina Panthers as the wide 
receivers coach. He continued to make significant contributions to the team's 
success, playing a vital role in the development of their receiving corps.
Known for his passion, knowledge, and dedication to the sport, McClendon has earned 
a stellar reputation within the football community. His ability to connect with 
players and maximize their potential has made him a highly sought-after coach. 
With his impressive resume and commitment to excellence, Bryan McClendon continues 
to leave a lasting impact in the world of football.
\end{verbatim}
\end{quote}

\subsection{SelfCheckGPT-Prompt (SCGP)} \label{sec:prompt_SCGP}

Function $supported(r_i, S^j)$ uses the below LLM prompt to do a binary determination of whether regenerated text $S^j$ supports original sentence $r_i$.
If the LLM completion is \textit{``Yes''}, the function returns 1.0; if \textit{``No''}, returns 0.0; otherwise, returns 0.5.

\tb{LLM parameters:}

\begin{itemize}
  \item model version = \emph{gpt-3.5-turbo-16k-0613}
  \item temperature = 0.0 (to eliminate stochastic fluctuations)
  \item top\_p = 0.0
\end{itemize}

\tb{Prompt Template:}

\begin{quote}
\begin{verbatim}
You are a helpful assistant.
---------------------------------------------------------------
Context: {context}

Sentence: {sentence}

Is the sentence supported by the context above? Answer Yes or No:
\end{verbatim}
\end{quote}

Where,
\begin{itemize}
  \item \verb|{context}| is replaced by regenerated text $S^j$
  \item \verb|{sentence}| is replaced by original sentence $r_i$
\end{itemize}

\tb{Prompt Example:}

\begin{quote}
\begin{verbatim}
You are a helpful assistant.
---------------------------------------------------------------
Context: Bryan McClendon is an accomplished American football coach and former 
player. Born on May 27, 1984, in Atlanta, Georgia, McClendon has an impressive 
career in the sport. He played as a wide receiver for the University of Georgia 
Bulldogs from 2002 to 2005, earning recognition for his exceptional skills and 
leadership on the field. Following his playing career, McClendon embarked on a 
successful coaching journey. He held various coaching positions, including 
serving as the wide receivers coach for the University of Georgia from 2009 to 
2015. During this time, he played a crucial role in developing talented receivers 
and contributing to the team's success. In 2015, McClendon was named the interim 
head coach for the Georgia Bulldogs, becoming the youngest head coach in the 
Southeastern Conference at just 31 years old. Despite only leading the team for 
one game, he showcased his ability to inspire and motivate the players. 
McClendon's coaching prowess caught the attention of the National Football League 
(NFL) teams, and in 2016, he joined the staff of the Carolina Panthers as the wide 
receivers coach. He continued to make significant contributions to the team's 
success, playing a vital role in the development of their receiving corps.  
Known for his passion, knowledge, and dedication to the sport, McClendon has 
earned a stellar reputation within the football community. His ability to connect 
with players and maximize their potential has made him a highly sought-after coach. 
With his impressive resume and commitment to excellence, Bryan McClendon continues 
to leave a lasting impact in the world of football.

Sentence: In 2012, he returned to Georgia as the running backs coach.

Is the sentence supported by the context above? Answer Yes or No:
\end{verbatim}
\end{quote}

\tb{Completion Example:}

\begin{quote}
\begin{verbatim}
No
\end{verbatim}
\end{quote}

\subsection{Multiple-Fill-in-the-Blank exam (FIBE)} \label{sec:prompt_FIBE}

\subsubsection{$create(R, P)$} \label{sec:prompt_FIBE_create}

Function $create(R, P)$ uses the below LLM prompt to extract words (objects) from original text $R$ to be fill-in-the-blank questions.
The function replaces the extracted objects with variable names based on their hypernyms extracted together, such as \textit{``\$year\_20''}, to create a fill-in-the-blank exam $E$.
Here, the objects appearing before the subject extracted together in each sentence are not masked to prevent topic picking.

\tb{LLM parameters:}

\begin{itemize}
  \item model version = \emph{gpt-3.5-tu
  rbo-16k-0613}
  \item temperature = 0.0 (to eliminate stochastic fluctuations)
  \item top\_p = 0.0
  \item random seed = 0
\end{itemize}

\tb{Prompt Template:}

\begin{quote}
\begin{verbatim}
You are an expert in natural language processing for English, so you output your 
answer in English.

You are now going to make up "Fill-in-the-blank Questions" based on the "Texts" 
for testing students' understanding.
Be sure to follow the instructions in the "Precautions" section. 
---------------------------------------------------------------
Making up "Fill-in-the-blank Questions" from only sentences with serial numbers 
in the following "Texts".

# Precautions
* Extract a subject of each sentence.
* Extract only single concrete eigenexpression as an blank; i.e., extract time, 
date, location, number, and, proper noun.
  + Select only few words as an object from a phrase containing three or more words; 
e.g., phrase "pathophysiology of many diseases" --> blank 
<pathophysiology:academic_field>.
  + Don't extract blanks that do not settle on one correct answer., such as 
"beautiful", "good", etc.
* Specify the hypernym as hint of each blank and subject;
  + e.g., <John Smith:person>, <31:day>, <July:month>, <2023:year>, 
<New York:city>, <Kanagawa:prefecture>, <World Cup:sports event>, 
<carpenter:profession>, <4:number of cars>
---------------------------------------------------------------
Texts:
``What kind of person is Alice?''
[s0] Alice Liddell (21 March 1955 - 1 Dec. 2020) is the founder of Philz.
[s1] Her branches were located in the USA and in Japan, for a total of two branches.
---------------------------------------------------------------
Fill-in-the-blank Questions: (from [s0],[s1])
Text=[s0] Alice Liddell (21 March 1955 - 1 Dec. 2020) is the founder of Philz.
Subject=<Alice Liddell:person>
Blanks=<21:day>, <March:month>, <1955:year>, <1:day>, <Dec.:month>, <2020:year>, 
<Philz:shop>
----
Text=[s1] Her branches were located in the USA and in Japan, for a total of two 
branches.
Subject=<Her branches:branches>
Blanks=<USA:country>, <Japan:country>, <two:number of branches>
---------------------------------------------------------------
Texts:
{context}{sentences}
---------------------------------------------------------------
Fill-in-the-blank Questions: (from {sids})
\end{verbatim}
\end{quote}

Where,
\begin{itemize}
  \item \verb|{context}| is replaced by original prompt $P$
  \item \verb|{sentences}| is replaced by the sentences, with their serial numbers, in original text $R$
  \item \verb|{sids}| is replaced by the serial numbers of the sentences
\end{itemize}

\tb{Prompt Example:}

\begin{quote}
\begin{verbatim}
You are an expert in natural language processing for English, so you output your 
answer in English.

You are now going to make up "Fill-in-the-blank Questions" based on the "Texts" 
for testing students' understanding.
Be sure to follow the instructions in the "Precautions" section. 
---------------------------------------------------------------
Making up "Fill-in-the-blank Questions" from only sentences with serial numbers 
in the following "Texts".

# Precautions
* Extract a subject of each sentence.
* Extract only single concrete eigenexpression as an blank; i.e., extract time, 
date, location, number, and, proper noun.
  + Select only few words as an object from a phrase containing three or more words; 
e.g., phrase "pathophysiology of many diseases" --> blank 
<pathophysiology:academic_field>.
  + Don't extract blanks that do not settle on one correct answer., such as 
"beautiful", "good", etc.
* Specify the hypernym as hint of each blank and subject;
  + e.g., <John Smith:person>, <31:day>, <July:month>, <2023:year>, 
<New York:city>, <Kanagawa:prefecture>, <World Cup:sports event>, 
<carpenter:profession>, <4:number of cars>
---------------------------------------------------------------
Texts:
``What kind of person is Alice?''
[s0] Alice Liddell (21 March 1955 - 1 Dec. 2020) is the founder of Philz.
[s1] Her branches were located in the USA and in Japan, for a total of two branches.
---------------------------------------------------------------
Fill-in-the-blank Questions: (from [s0],[s1])
Text=[s0] Alice Liddell (21 March 1955 - 1 Dec. 2020) is the founder of Philz.
Subject=<Alice Liddell:person>
Blanks=<21:day>, <March:month>, <1955:year>, <1:day>, <Dec.:month>, <2020:year>, 
<Philz:shop>
----
Text=[s1] Her branches were located in the USA and in Japan, for a total of two 
branches.
Subject=<Her branches:branches>
Blanks=<USA:country>, <Japan:country>, <two:number of branches>
---------------------------------------------------------------
Texts:
``This is a Wikipedia passage about Bryan McClendon:''
[s0] Bryan McClendon (born April 28, 1983) is an American football coach and former 
player.
[s1] He is currently the offensive coordinator and quarterbacks coach at the 
University of Georgia.
[s2] He previously served as the co-offensive coordinator and running backs coach 
at the University of South Carolina.
[s3] McClendon played college football at the University of Georgia, where he was 
a four-year letterman and two-year starter at wide receiver.
[s4] He was a member of the Bulldogs' 2002 SEC Championship team and the 2005 Sugar 
Bowl team.
[s5] After his playing career, McClendon served as a graduate assistant at Georgia 
from 2006 to 2008.
[s6] He then served as the wide receivers coach at the University of Tennessee from 
2009 to 2011.
[s7] In 2012, he returned to Georgia as the running backs coach.
[s8] In 2014, he was promoted to co-offensive coordinator and running backs coach.
[s9] In 2016, he left Georgia to become the co-offensive coordinator and running 
backs coach at South Carolina.
[s10] In 2018, he returned to Georgia as the offensive coordinator and quarterbacks 
coach.
---------------------------------------------------------------
Fill-in-the-blank Questions: (from [s0],[s1],[s2],[s3],[s4],[s5],[s6],[s7],[s8],
[s9],[s10])
\end{verbatim}
\end{quote}

\tb{Completion Example:}

\begin{quote}
\begin{verbatim}
Text=[s0] Bryan McClendon (born April 28, 1983) is an American football coach and 
former player.
Subject=<Bryan McClendon:person>
Blanks=<April:month>, <28:day>, <1983:year>, <American:nationality>, 
<football coach:profession>, <former player:profession>
----
Text=[s1] He is currently the offensive coordinator and quarterbacks coach at the 
University of Georgia.
Subject=<He:person>
Blanks=<offensive coordinator:profession>, <quarterbacks coach:profession>, 
<University of Georgia:university>
----
Text=[s2] He previously served as the co-offensive coordinator and running backs 
coach at the University of South Carolina.
Subject=<He:person>
Blanks=<co-offensive coordinator:profession>, <running backs coach:profession>, 
<University of South Carolina:university>
----
Text=[s3] McClendon played college football at the University of Georgia, where 
he was a four-year letterman and two-year starter at wide receiver.
Subject=<McClendon:person>
Blanks=<college football:activity>, <University of Georgia:university>, 
<four-year:number of years>, <letterman:position>, <two-year:number of years>, 
<wide receiver:position>
----
Text=[s4] He was a member of the Bulldogs' 2002 SEC Championship team and the 2005 
Sugar Bowl team.
Subject=<He:person>
Blanks=<Bulldogs:team>, <2002:year>, <SEC Championship:championship>, <2005:year>, 
<Sugar Bowl:game>
----
Text=[s5] After his playing career, McClendon served as a graduate assistant at 
Georgia from 2006 to 2008.
Subject=<McClendon:person>
Blanks=<graduate assistant:profession>, <Georgia:university>, <2006:year>, 
<2008:year>
----
Text=[s6] He then served as the wide receivers coach at the University of Tennessee 
from 2009 to 2011.
Subject=<He:person>
Blanks=<wide receivers coach:profession>, <University of Tennessee:university>, 
<2009:year>, <2011:year>
----
Text=[s7] In 2012, he returned to Georgia as the running backs coach.
Subject=<he:person>
Blanks=<2012:year>, <Georgia:university>, <running backs coach:profession>
----
Text=[s8] In 2014, he was promoted to co-offensive coordinator and running backs 
coach.
Subject=<he:person>
Blanks=<2014:year>, <co-offensive coordinator:profession>, 
<running backs coach:profession>
----
Text=[s9] In 2016, he left Georgia to become the co-offensive coordinator and 
running backs coach at South Carolina.
Subject=<he:person>
Blanks=<2016:year>, <Georgia:university>, <co-offensive coordinator:profession>, 
<running backs coach:profession>, <South Carolina:university>
----
Text=[s10] In 2018, he returned to Georgia as the offensive coordinator and 
quarterbacks coach.
Subject=<he:person>
Blanks=<2018:year>, <Georgia:university>, <offensive coordinator:profession>, 
<quarterbacks coach:profession>
\end{verbatim}
\end{quote}

\subsubsection{$answer(E, P)$} \label{sec:prompt_FIBE_answer}

Function $answer(E, P)$ uses the below LLM prompt to fill in the blanks in the exam $E$.
This prompt is batchable, and the number of simultaneous completions can be set by the parameter \emph{n} in \emph{GPT-3.5}.

\tb{LLM parameters:}

\begin{itemize}
  \item model version = \emph{gpt-3.5-turbo-16k-0613}
  \item temperature = 0.5 (to ensure successful instruction following)
  \item top\_p = 1.0
  \item random seed = 0
\end{itemize}

\tb{Prompt Template:}

\begin{quote}
\begin{verbatim}
You are the world champion in English quizzes.

You are now going to answer the "Fill-in-the-blank Questions".
Be sure to follow the instructions in the "Precautions" section.
Be sure to output the serial number of each sentence (e.g., "[s0]", "[s3]").
---------------------------------------------------------------
Output new "Fill-in-the-blank Answers" that fills in the variables in the 
following "Fill-in-the-blank Questions" with concrete variable values.

# Precautions
* The variable naming convention is "$HINT_NUMBER"; e.g., "$date_0".
* Each variable value has a different value each other.
* Terms that are not variables in each sentence should be left as they are.
---------------------------------------------------------------
Fill-in-the-blank Questions:
``What kind of person is Alice?''
[s0] Alice (born "$date_0") is the founder of "$place_1".
[s1] It is a "$place_2" founded in "$location_3" in "$year_4".
---------------------------------------------------------------
Fill-in-the-blank Answers: (up to [s1])
``What kind of person is Alice?''
[s0] Alice (born 21 March 1955) is the founder of Philz.
[s1] It is a coffee shop founded in Berkeley in 1985.
---------------------------------------------------------------
Fill-in-the-blank Questions:
{context}{source}
---------------------------------------------------------------
Fill-in-the-blank Answers: (up to [s{max_sentences}])
{context}
\end{verbatim}
\end{quote}

Where,
\begin{itemize}
  \item \verb|{context}| is replaced by original prompt $P$
  \item \verb|{source}| is replaced by the sentences, with their serial numbers, in fill-in-the-blank exam $E$
  \item \verb|{max_sentences}| is replaced by the largest serial number out of the sentences
\end{itemize}

\tb{Prompt Example:}

\begin{quote}
\begin{verbatim}
You are the world champion in English quizzes.

You are now going to answer the "Fill-in-the-blank Questions".
Be sure to follow the instructions in the "Precautions" section.
Be sure to output the serial number of each sentence (e.g., "[s0]", "[s3]").
---------------------------------------------------------------
Output new "Fill-in-the-blank Answers" that fills in the variables in the 
following "Fill-in-the-blank Questions" with concrete variable values.

# Precautions
* The variable naming convention is "$HINT_NUMBER"; e.g., "$date_0".
* Each variable value has a different value each other.
* Terms that are not variables in each sentence should be left as they are.
---------------------------------------------------------------
Fill-in-the-blank Questions:
``What kind of person is Alice?''
[s0] Alice (born "$date_0") is the founder of "$place_1".
[s1] It is a "$place_2" founded in "$location_3" in "$year_4".
---------------------------------------------------------------
Fill-in-the-blank Answers: (up to [s1])
``What kind of person is Alice?''
[s0] Alice (born 21 March 1955) is the founder of Philz.
[s1] It is a coffee shop founded in Berkeley in 1985.
---------------------------------------------------------------
Fill-in-the-blank Questions:
``This is a Wikipedia passage about Bryan McClendon:''
[s0] Bryan McClendon (born "$month_3" "$day_4", "$year_0") is an "$nationality_5" 
"$profession_1" and "$profession_2".
[s1] He is currently the "$profession_6" and "$profession_7" at the 
"$university_8".
[s2] He previously served as the "$profession_9" and "$profession_11" at the 
"$university_10".
[s3] McClendon played "$activity_12" at the "$university_8", where he was a 
"$number of years_13" "$position_14" and "$number of years_15" starter at 
"$position_16".
[s4] He was a member of the "$team_17"' "$year_18" "$championship_19" team and the 
"$year_20" "$game_21" team.
[s5] After his playing career, McClendon served as a "$profession_22" at 
"$university_25" from "$year_23" to "$year_24".
[s6] He then served as the "$profession_26" at the "$university_27" from 
"$year_28" to "$year_29".
[s7] In 2012, he returned to "$university_25" as the "$profession_11".
[s8] In 2014, he was promoted to "$profession_9" and "$profession_11".
[s9] In 2016, he left "$university_25" to become the "$profession_9" and 
"$profession_11" at "$university_33".
[s10] In 2018, he returned to "$university_25" as the "$profession_6" and 
"$profession_7".
---------------------------------------------------------------
Fill-in-the-blank Answers: (up to [s10])
\end{verbatim}
\end{quote}

\tb{Completion Example:}

\begin{quote}
\begin{verbatim}
[s0] Bryan McClendon (born December 28, 1982) is an American football coach and 
former player.
[s1] He is currently the wide receivers coach and passing game coordinator at the 
University of Oregon.
[s2] He previously served as the wide receivers coach and co-offensive coordinator 
at the University of Georgia.
[s3] McClendon played college football at the University of Georgia, where he was a
four-year wide receiver and two-year starter at quarterback.
[s4] He was a member of the 2002 SEC Championship team and the 2003 Sugar Bowl team.
[s5] After his playing career, McClendon served as a graduate assistant at the 
University of Georgia from 2006 to 2008.
[s6] He then served as the wide receivers coach at the University of Georgia from 
2009 to 2015.
[s7] In 2012, he returned to the University of Georgia as the running backs coach.
[s8] In 2014, he was promoted to wide receivers coach and co-offensive coordinator.
[s9] In 2016, he left the University of Georgia to become the wide receivers coach 
and co-offensive coordinator at the University of South Carolina.
[s10] In 2018, he returned to the University of Georgia as the wide receivers coach 
and passing game coordinator.
\end{verbatim}
\end{quote}

\subsubsection{$score(a^j_i, r_i)$} \label{sec:prompt_FIBE_score}

Function $score(a^j_i, r_i)$ uses the below LLM prompt to score the consistency of original sentence $r_i$ with answer $a^j_i$ on a 100-point scale.
This prompt can also score multiple examinees' answers $a^{1 \le j \le N}_i$ together.

\tb{LLM parameters:}

\begin{itemize}
  \item model version = \emph{gpt-3.5-turbo-16k-0613}
  \item temperature = 0.0 (to eliminate stochastic fluctuations)
  \item top\_p = 0.0
  \item random seed = 0
\end{itemize}

\tb{Prompt Template:}

\begin{quote}
\begin{verbatim}
You are an English test grader.

A student's answer to a fill-in-the-blank question should be scored between 
0 and 100 points based on a comparison with the "Correct answer".
  * A score of 0 shall be scored if the student answers is in complete contradiction 
with the "Correct answer"
  * A score of 100 shall be scored if the student answers is in complete agreement 
with the "Correct answer"
  * Score them very carefully, as you only want to pass the very best students.
---------------------------------------------------------------
Correct answer: {correct_answer}

Student name (inside square brackets) and answers:
{student_answers}

Answer only the name and score for each student, such as "[Sat] 50":
\end{verbatim}
\end{quote}

Where,
\begin{itemize}
  \item \verb|{correct_answer}| is replaced by original sentence $r_i$
  \item \verb|{student_answers}| is replaced by the answers $a^{1 \le j \le N}_i$ with the predefined examinee's names, such as \textit{``Tom''}.
\end{itemize}

\tb{Prompt Example:}

\begin{quote}
\begin{verbatim}
You are an English test grader.

A student's answer to a fill-in-the-blank question should be scored between 
0 and 100 points based on a comparison with the "Correct answer".
  * A score of 0 shall be scored if the student answers is in complete contradiction 
with the "Correct answer"
  * A score of 100 shall be scored if the student answers is in complete agreement 
with the "Correct answer"
  * Score them very carefully, as you only want to pass the very best students.
---------------------------------------------------------------
Correct answer: In 2012, he returned to Georgia as the running backs coach.

Student name (inside square brackets) and answers:
[Tom] In 2012, he returned to Georgia as the Wide Receivers Coach.
[Amy] In 2012, he returned to Georgia as the running backs coach.
[Max] In 2012, he returned to Georgia as the running backs coach.
[Leo] In 2012, he returned to the University of Georgia as the running backs coach.
[Ava] In 2012, he returned to the University of Georgia as the wide receivers coach.

Answer only the name and score for each student, such as "[Sat] 50":
\end{verbatim}
\end{quote}

\tb{Completion Example:}

\begin{quote}
\begin{verbatim}
[Tom] 0
[Amy] 100
[Max] 100
[Leo] 75
[Ava] 0
\end{verbatim}
\end{quote}

\subsection{Direct Question (DQ)} \label{sec:prompt_DQ}

Function $known(r_i, P)$ uses the below LLM prompt to do a binary determination of whether the prior knowledge of the LLM supports original sentence $r_i$ under context $P$.
If the LLM completion is \textit{``Yes''}, the function returns 1.0; if \textit{``No''}, returns 0.0; otherwise, returns 0.0.
Additionally, if the LLM output contains each occurrence probability $p_k \in [0,1]$ of the $k$-th output token as \emph{GPT-3.5} does, the function returns $p_1$ in case of \textit{``Yes''}, returns $1 - p_1$ in case of \textit{``No''}; otherwise, returns 0.0.

\tb{LLM parameters:}

\begin{itemize}
  \item model version = \emph{gpt-3.5-turbo-16k-0613}
  \item temperature = 0.0 (to eliminate stochastic fluctuations)
  \item top\_p = 0.0
  \item random seed = 0
\end{itemize}

\tb{Prompt Template:}

\begin{quote}
\begin{verbatim}
You are a machine-learning model that responds using only your prior knowledge.
---------------------------------------------------------------
{context}
Claim:{claim}
====
Is the above claim true?
Answer only Yes or No:
\end{verbatim}
\end{quote}

Where,
\begin{itemize}
  \item \verb|{context}| is replaced by original prompt $P$
  \item \verb|{claim}| is replaced by original sentence $r_i$
\end{itemize}

\tb{Prompt Example:}

\begin{quote}
\begin{verbatim}
You are a machine-learning model that responds using only your prior knowledge.
---------------------------------------------------------------
``This is a Wikipedia passage about Bryan McClendon:''
Claim:In 2012, he returned to Georgia as the running backs coach.
====
Is the above claim true?
Answer only Yes or No:
\end{verbatim}
\end{quote}

\tb{Completion Example:}

\begin{quote}
\begin{verbatim}
Yes
\end{verbatim}
\end{quote}

\section{Hallucination Snowballing Example} \label{sec:example_snowballing}

Although \textit{``Stan Heal''} is a well-known Australian rules footballer \footnote{\url{https://en.wikipedia.org/wiki/Stan_Heal}}, LLM outputted \textit{``American former professional basketball player''} in the first sentence, in consequence, fabricated all subsequent sentences as hallucinations.

\tb{Original Prompt:}

\begin{quote}
\begin{verbatim}
You are the author of an excellent Wikipedia article. 
---------------------------------------------------------------
This is a Wikipedia passage about Stan Heal:
\end{verbatim}
\end{quote}

\tb{Original Text with Snowballing:}

\begin{quote}
\begin{verbatim}
Stan Heal (born October 28, 1932) is an American former professional basketball 
player. He played in the National Basketball Association (NBA) from 1954 to 1965 
as a member of the Minneapolis Lakers, St. Louis Hawks, and Detroit Pistons.

Heal was a 6'3" guard from the University of Minnesota. He was selected by the 
Minneapolis Lakers in the 1954 NBA draft. He played in the NBA for 11 seasons, 
averaging 8.2 points per game. He was a member of the Lakers' 1959 NBA championship
team. He was also a member of the St. Louis Hawks' 1958 NBA championship team.

Heal was known for his defensive prowess and was selected to the NBA All-Defensive
Second Team in 1962. He was also selected to the NBA All-Star Game in 1959. 
After retiring from the NBA, Heal coached the Detroit Pistons for two seasons. 
He was inducted into the Minnesota Basketball Hall of Fame in 1994.
\end{verbatim}
\end{quote}

\end{document}